\documentclass[12pt,a4paper]{article}

\usepackage{lec}
\usepackage{color}
\usepackage{geometry}

\usepackage{amsfonts}
\usepackage{amsthm}
\usepackage{graphicx}
\usepackage{url}

\usepackage{hyperref}
\usepackage{multicol}
\usepackage{lipsum}

\usepackage{amsmath}
\usepackage{bm}

\title{Example Perplexity\footnote{Communications: {\tt lzhang@cse.ust.hk}, {\tt lixiaohui33@huawei.com}}}

\author{\normalsize Nevin L. Zhang$^{1}$,  Weiyan Xie$^{1}$, Zhi Lin$^{1}$, Guanfang Dong$^{2}$\\
\normalsize Xiao-Hui Li$^{3}$, Caleb Chen Cao$^{3}$, Yunpeng Wang$^{3}$ \\
\normalsize $^{1}$The Hong Kong University of Science and Technology \\
\normalsize $^{2}$University of Alberta \\
 \normalsize $^{3}$Huawei Technologies Co., Ltd}

\date{}

\begin{document}
\maketitle

\begin{abstract}

Some examples are easier for humans to classify than others. The same should be true for deep neural networks (DNNs). We use the term {\em example perplexity} to refer to the level of difficulty of classifying an example.
In this paper, we propose a method to measure the perplexity of an example
and investigate what factors contribute to high example perplexity. The related codes and resources are available at \url{https://github.com/vaynexie/Example-Perplexity}.

\end{abstract}

\section{Introduction}

Some examples are easier for humans to classify than others. The same should be true for deep neural networks (DNNs). We use the term example perplexity to refer to the level of difficulty of classifying an example.
While scarce, there are previous works that consider classification difficulty of examples and datasets.
Ionescu {\em et al}. \cite{tudor2016hard} estimate  {\em image difficulty}  by learning a regression model from human-originated scores, which are converted from  response times during visual search tasks.
Yu {\em et al}. \cite{yu2020difficulty} view disagreement among experts when labeling medical images as an indication of their  difficulty.  Both of those two methods rely on human annotations and hence do not scale up.
Scheidegger {\em et al}.
\cite{scheidegger2020efficient} use the average performance of top  models as a measure of the {\em difficulty of a dataset}.
Li {\em et al}.\cite{li2017not} and Nie {\em et al}.
\cite{nie2019difficulty} exploit {\em pixel difficulty} in the context of semantic segmentation.  Those two methods are not about example difficulty.

In this paper, we propose a method to measure the perplexity of an example
and investigate what factors contribute to high example perplexity.
 To estimate the perplexity of an image for DNNs, we create a population of DNN classifiers with varying architectures and trained on data of varying sample sizes, just as different people have different IQs and different amounts of experiences. For an unlabeled example, the average entropy of the output probability distributions of the classifiers is taken to be the C-perplexity of the example, where C stands for ``confusion”. For a labeled example, the fraction of classifiers that misclassify the example is taken to be the X-perplexity of the example, where X stands for ``mistake”.

Perplexity analysis on the examples from ImageNet has revealed
several interesting. In particular, some insights are gained regarding
what makes some images harder to classify than others.
It is also found that perplexity analysis can reveal imperfections of a dataset, which can potentially help with data cleaning

While we use a large population of classifiers consisting of both strong and week models to measure example perplexity,
Tsipras {\em et al}. \cite{tsipras2020imagenet} and Beyer {\em et al.} \cite{beyer2020we} use a small population of state-of-the-art classifiers to suggest candidate labels for images for the purpose of data quality improvement.

\section{C-Perplexity and X-Perplexity}

An image is difficult to classify for humans if many people find it confusing and classify it incorrectly.  Motivated by this observation, we measure example perplexity in reference to a population classifiers.
Let $\sC$ be a population of $N$ classifiers for classifying examples into $M$ classes.
For a given example $\x$, $P_i(y|\x)$ is the probability distribution over the $M$ classes computed by classifier $i$.  The entropy $H(P_i(y|\x)) = - \sum_y P_i(y|\x) \log_2 P_i(y|\x)$  is a measure of how uncertain the classifier is when classifying $\x$.  The {\em perplexity of the probability distribution} is defined to be $2^{H(P_i(y|\x))}$ \footnote{\url{https://en.wikipedia.org/wiki/Perplexity}}.
The larger the perplexity, the less confident the classifier is about its prediction. When the distribution places equal probability on $k$ possible classes and zero probability on others, the perplexity is $k$.

We define the {\em C-perplexity}  of an unlabelled example $\x$  w.r.t $\sC$ to be the following geometric mean:

\[\Phi_{C}(\x) =  [\prod_{i=1}^N 2^{ H(P_i(y|\x))}]^{\frac{1}{N}}.\]

\noindent The prefix ``C" stands for ``confusion".  The minimum possible value of C-perplexity is 1. High C-perplexity value indicates that the classifiers have low confidence when classifying the example.

We define
the {\em X-perplexity} of an labelled example $(\x, y)$ w.r.t $\sC$ to be
\[\Phi_{X}(\x) = \frac{1}{N} \sum_{i=1}^N \1(C_i(\x) \neq y),\]
\noindent where
\(C_i(\x) = \arg \max_{y} P_i(y|\x)\) is the class assignment function,  and $\1$ is the indicator function.  In words, it is the fraction of the classifiers that misclassifies the example, hence is between 0 and 1.
 The prefix ``X" stands for ``misclassification".

\section{Creation of a Classifier Population}

We have created a population of 500 classifiers of varying strengths.  We started  with 10 popular model architectures (Table \ref{tab:seeds}) designed for the ImageNet dataset, which consists of approximately 14 million training examples.  Besides the original ImageNet training set, we created 9 smaller training sets via sub-sampling without replacement. They are evenly divided into three groups, with 25\%, 50\% and 75\% of the training examples respectively.  We trained each of the 10 architectures on each of 10 training sets. During each training session, four models were collected at different epochs prior and close to convergence, and one model was collected at convergence.

\begin{table}[h]
\centering
\caption{Seed model architectures for our classifier population.}
\label{tab:seeds}
\begin{tabular}{|c|c|c|}
\hline
Structure      & \# Parameters (M) & Storage Space (MB) \\ \hline
VGG16          & 138               & 500                \\ \hline
ResNet50       & 25                & 98                 \\ \hline
ResNet101      & 44                & 171                \\ \hline
InceptionV3    & 24                & 92                 \\ \hline
Xception       & 23                & 88                 \\ \hline
DenseNet121    & 8                 & 33                 \\ \hline
DenseNet169    & 14                & 57                 \\ \hline
DenseNet201    & 20                & 80                 \\ \hline
EfficientNetB0 & 5.3               & 20                 \\ \hline
EfficientNetB2 & 9                 & 36                 \\ \hline
\end{tabular}
\end{table}

Generally speaking,  classifiers trained with less data and for fewer epochs are weaker than those trained with more data and for more epochs.  Why do we need weak classifiers in addition to strong ones?  This question can be answered by making an analogy to educational testing, where a key concern is to design test items with appropriate difficult levels.  To determine the difficult level of a particular item, responses from a population of students need to be collected and analyzed using item response theory \cite{embretson2013item}.  It is evident that one cannot properly determine the difficult level of a test item by trying it only on strong students.  Similarly, we need classifiers of varying strengths to properly determine the classification difficulty of an image.  Architectures with different complexity can be compared to students with different IQs. Different training sets with various sample sizes can be compared to different experiences students have access to. Different training epochs can be compared to how much students digest their experiences.

Note that, while ensemble learning involves multiple classifiers that are complementary in strength, we need multiple classifiers with a variety of strengths in this project.

\section{Perplexity Analysis of the ImageNet DataSet}

Based on the aforementioned 500 classifiers, we have computed the C-perplexity and X-perplexity of the 50,000 images in the validation set of ImageNet. Figure \ref{fig.cp-examples} shows several examples with  C-perplexity values ranging from 1 to 15, and  Figure \ref{fig.xp-examples} shows several examples with  x-perplexity values ranging from 0 to 1.   In Section \ref{image-diff}, we will analyze the factors that contribute to high perplexity.

\begin{center}
\begin{figure}[h!]

\begin{tabular}{ccc}

\begin{tabular}{c}
 \includegraphics[width=4cm]{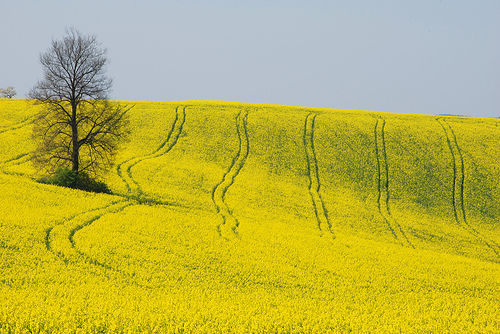}
\end{tabular}

&
\begin{tabular}{c}
 \includegraphics[width=4cm]{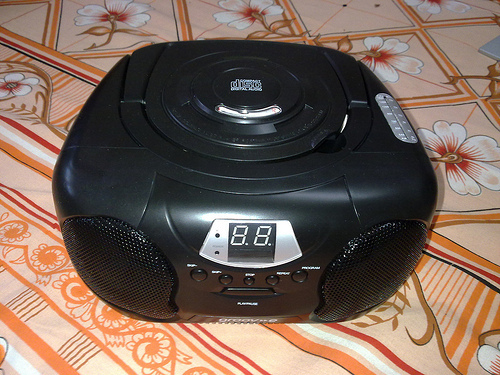}
\end{tabular}
&
\begin{tabular}{c}
 \includegraphics[width=4cm]{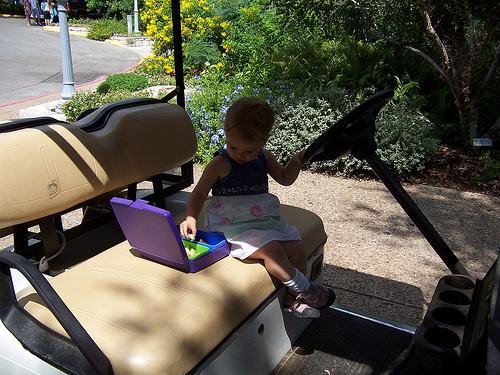}
\end{tabular}
\\
Label = {\tt \small rapeseed}  & Label = {\tt \small CD player}      &  Label = {\tt \small golfcart}\\
{\bf  CP = 1}                  &   {\bf  CP =2 }                   & {\bf  CP =3 } \\
XP = 0                  &  XP =0.15                     & XP =0.29 \\

\begin{tabular}{c}
 \includegraphics[width=2.7cm]{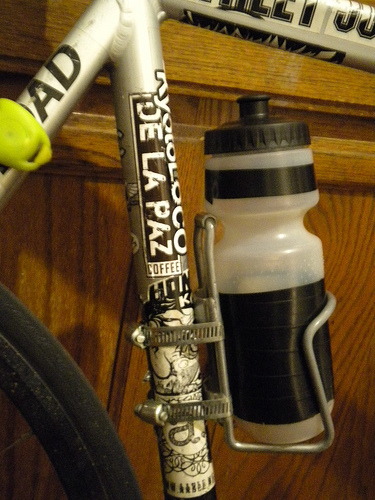}
\end{tabular}

&
\begin{tabular}{c}
 \includegraphics[width=4cm]{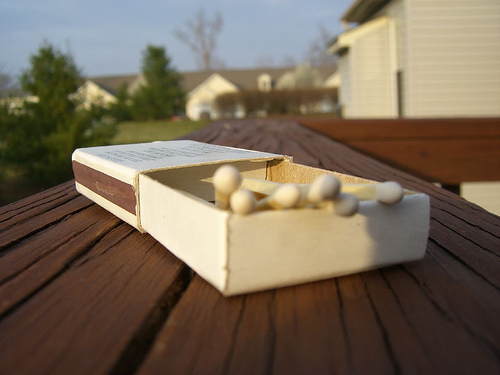}
\end{tabular}
&
\begin{tabular}{c}
 \includegraphics[width=4cm]{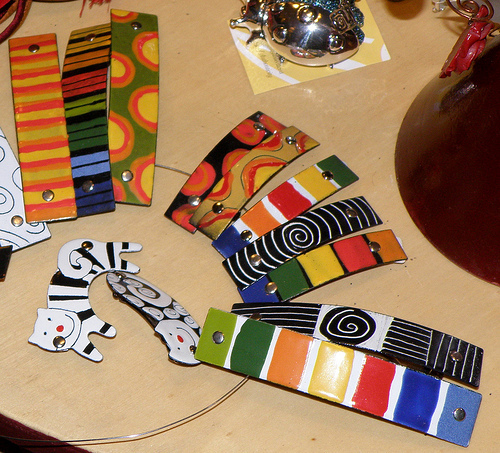}
\end{tabular}
\\
Label = {\tt \small water bottle } & Label = {\tt \small match stick}   &
Label = {\tt \small hair slide }\\
{\bf  CP = 5 }                 &  {\bf  CP = 10} &             {\bf  CP =15} \\
XP = 0.74                &  XP =0.99 &           XP =0.99 \\

\end{tabular}
%\vspace{-0.5cm}
\caption{Examples with varying C-perplexity values. }
\label{fig.cp-examples}
\end{figure}
\vspace{-1cm}
\end{center}

\begin{center}
\begin{figure}[h!]

\begin{tabular}{ccc}

\begin{tabular}{c}
 \includegraphics[width=2.5cm]{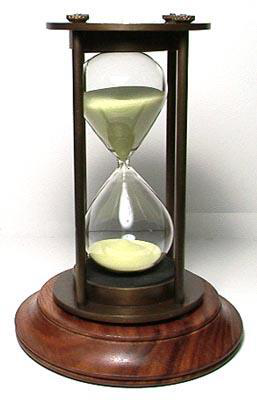}
\end{tabular}

&
\begin{tabular}{c}
 \includegraphics[width=4cm]{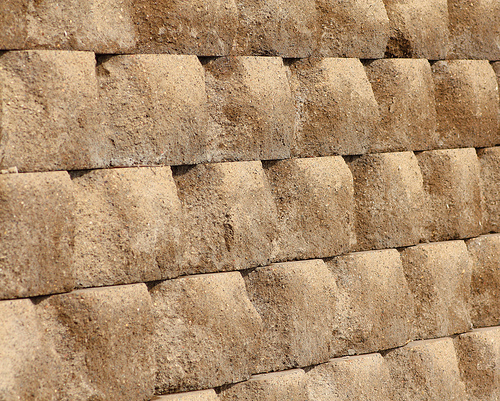}
\end{tabular}
&
\begin{tabular}{c}
 \includegraphics[width=4cm]{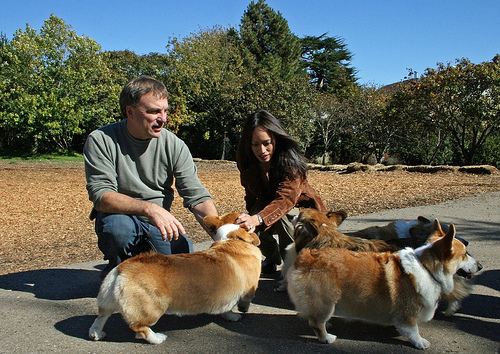}
\end{tabular}
\\
Label = {\tt \small hourglass}  & Label = {\tt \small stone wall}
 &  Label = {\tt \small mousetrap}\\
{\bf XP = 0 }                 &  {\bf  XP =0.2 }                   & {\bf  XP =0.4} \\
CP = 1                  &  CP =1.14                  & CP =8.07 \\

\begin{tabular}{c}
 \includegraphics[width=4cm]{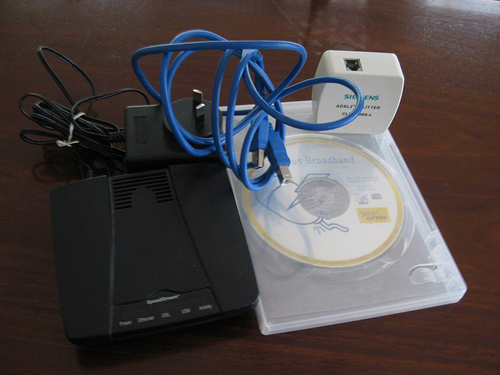}
\end{tabular}

&
\begin{tabular}{c}
 \includegraphics[width=4cm]{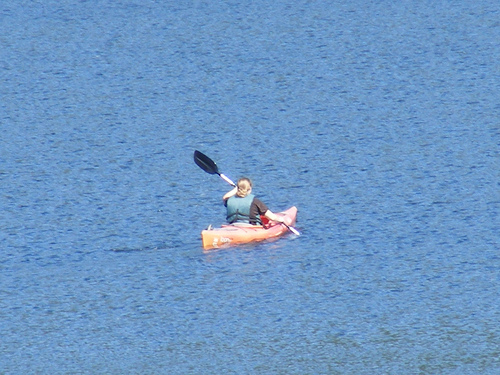}
\end{tabular}
&
\begin{tabular}{c}
 \includegraphics[width=4cm]{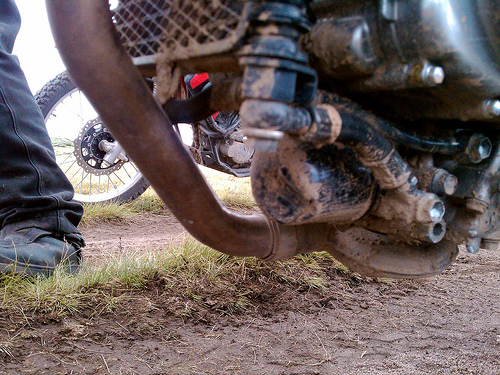}
\end{tabular}
\\
Label = {\tt \small modem } & Label = {\tt \small paddle}   &
Label = {\tt \small oil filter }\\
{\bf  XP = 0.6 }               & {\bf   XP =0.8} &          {\bf    XP =1.0} \\
CP = 4.37                &  CP = 10.9&             CP =14.7 \\

\end{tabular}
%\vspace{-0.5cm}
\caption{Examples with varying X-perplexity values. }
\label{fig.xp-examples}
\end{figure}
\vspace{-1cm}
\end{center}

\begin{figure}[h!]
\begin{tabular}{cc}
\begin{tabular}{c}
\hspace{-1cm} \includegraphics[width=12cm]{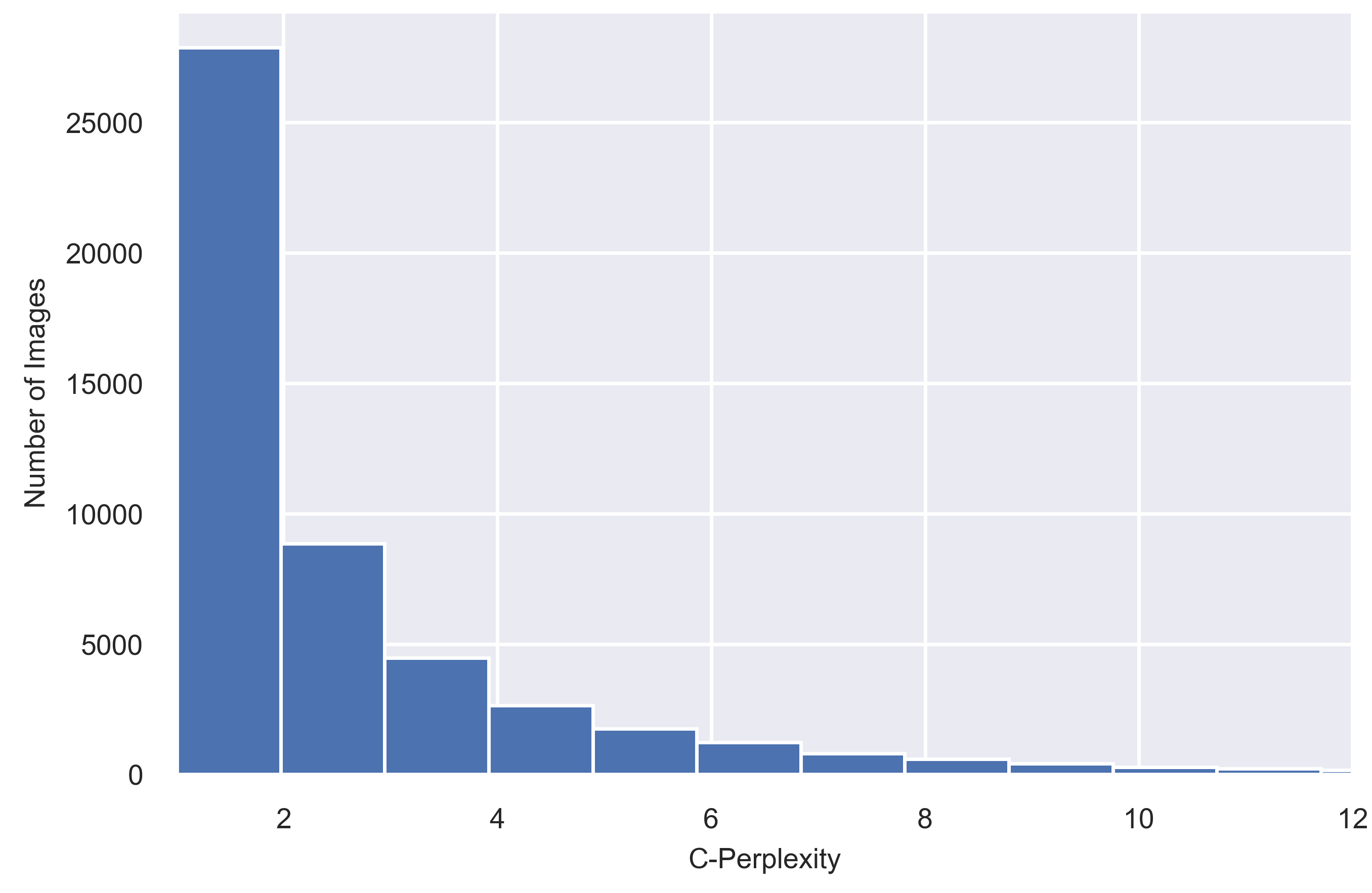}
\end{tabular}

 &
\begin{tabular}{c}
\hspace{-1cm} \includegraphics[width=4cm]{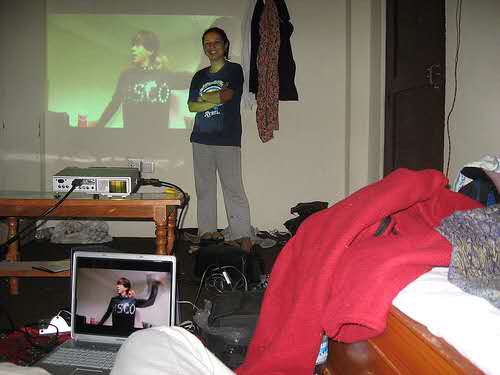}\\
\hspace{-1cm} Label = {\tt \small home theatre }
\end{tabular}
\end{tabular}
\caption{Distribution of examples over C-perplexity values (left), and one image with C-Perplexity value larger than 30 (right).}
\label{cp-distribution}
\vspace{-1cm}
\end{figure}

Figure \ref{cp-distribution} shows the distribution of the examples over the  C-perplexity values.  As mentioned earlier, a C-perplexity value of $k$ amounts to the uncertainty of choosing among $k$ equally probable classes.  The distribution indicates that, in most cases, the classifiers can narrow down the number of possible classes to  small numbers. However, there are 5 images with C-perplexity larger than 30. One example is given in the figure.

\bc
\ec

\begin{figure}[h!]
\includegraphics[width=15cm]{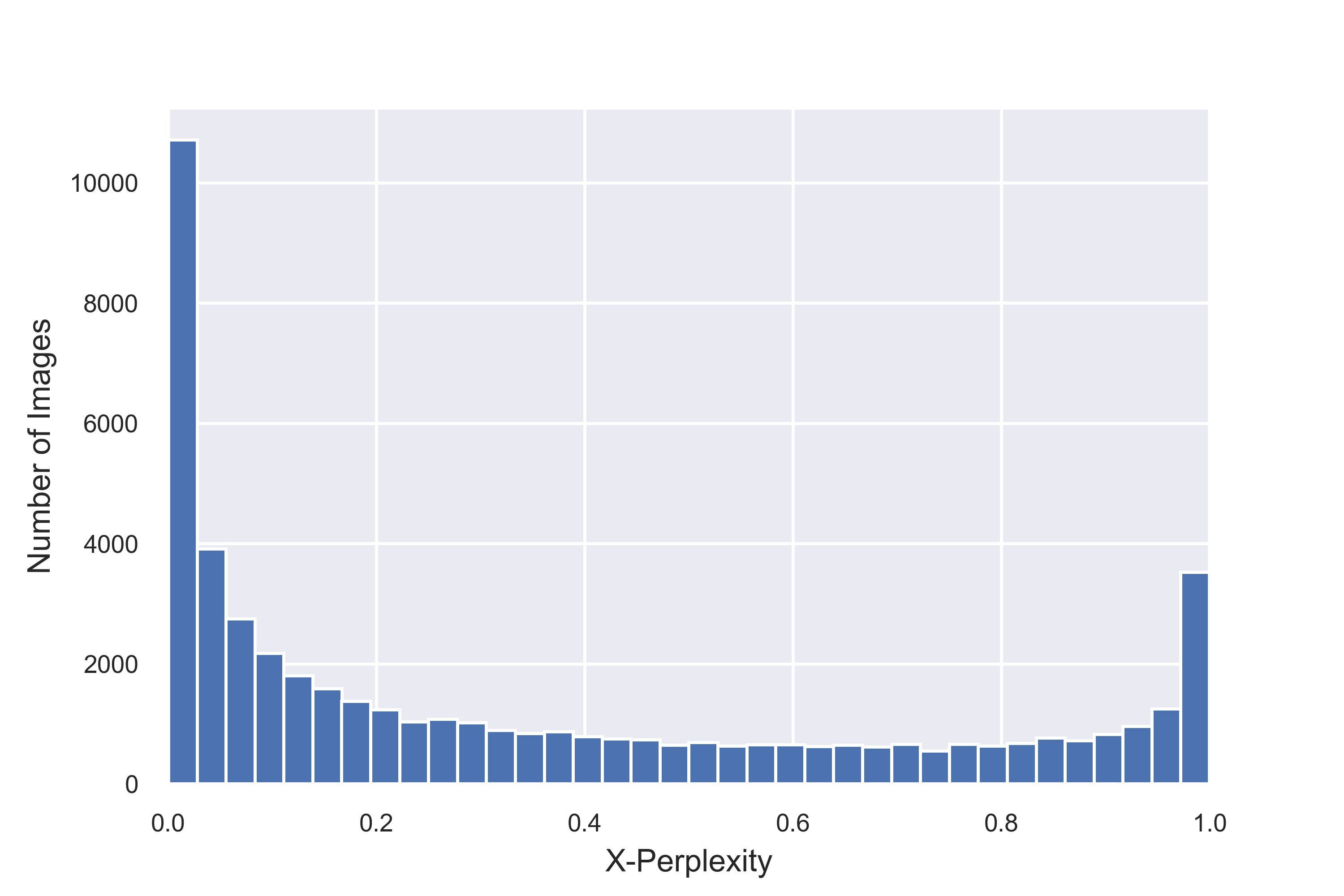}
\caption{Distribution of examples over X-perplexity values.   }
\label{xp-distribution}
\vspace{-1cm}
\end{figure}

Figure \ref{xp-distribution} shows the distribution of the example over the X-perplexity values. It is interesting to observe that the curve is U-shaped. This means that, while the majority of the examples are easy to classify, many of them are difficult for most and even all the classifiers.  There are two reasons: First,  some images are genuinely confusing to the classifiers (Section \ref{image-diff}). Second, some images are incorrectly or inappropriately labeled in the ImageNet dataset (Section \ref{imperfections}).  The two reasons will be analyzed in details later in this paper.

\bc
\ec

C-perplexity and X-perplexity are strongly correlated.  They are both influenced by the population of classifiers that is used in the computation. Detailed discussions of those issues can be found in Appendices 1 and 2 respectively.

\section{What Makes an Image Difficult to Classify?}
\label{image-diff}

There are multiple factors that make an image difficult to classify, for example,   clutter, occlusions, and adversarial perturbations.
An inspection of examples with high perplexity reveals two other conceptually interesting reasons that we will exploit later in this project.  The first one is  {\em attention confusion}, which refers to the difficulty that a classifier faces, when classifying an image with multiple objects, in deciding which object to place attention on. The second one is
{\em class confusion}, which refers to the difficulty that  a classifier faces  in deciding, among multiple visually similar  classes,  which class an object belongs to.  The two concepts are illustrated in Figure \ref{fig.confusion}.

In the case of attention confusion, a classifier is uncertain about which object to focus on and hence outputs significant probabilities for multiple class labels, with different labels corresponding to different regions of the image. This leads to high C-perplexity.  In addition, no matter which class is designated as the label of the image, some classifiers will  assign the image to other competing classes, resulting in high X-perplexity.
In the case of class confusion, a  classifier typically outputs significant probabilities for multiple visually similar classes, leading to high C-perplexity.  In addition, classifiers might assign the image to the wrong classes, resulting in high X-perplexity.

\begin{center}
\begin{figure}[t]
\begin{tabular}{cc}
\begin{tabular}{c}  %left
\begin{tabular}{c}
\includegraphics[width=4.5cm]{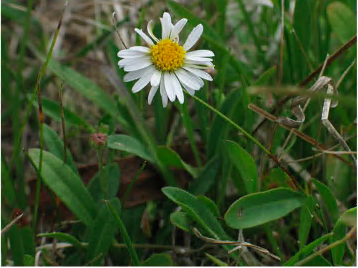} \\
(a) {\small Low perplexity}
\end{tabular} \\

\begin{tabular}{c}
\includegraphics[width=4.5cm]{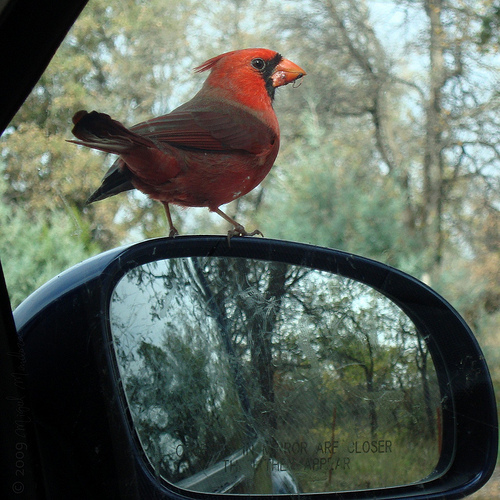} \\
(b) {\small Attention confusion}
\end{tabular}

\end{tabular}
\begin{tabular}{c}  %right
\begin{tabular}{cc}
\begin{tabular}{c}
\includegraphics[width=4.5cm]{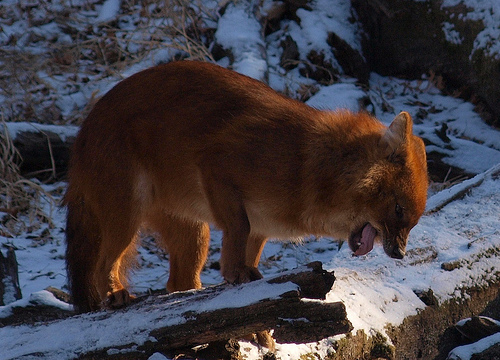}
\end{tabular}
&
\begin{tabular}{c}
\includegraphics[width=4.5cm]{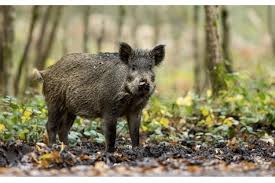}
\end{tabular}  \\
{\small \em True label: Dhole}  &   {\small \em Similar class:  Wild boar} \\
\begin{tabular}{c}
 \includegraphics[width=4.5cm]{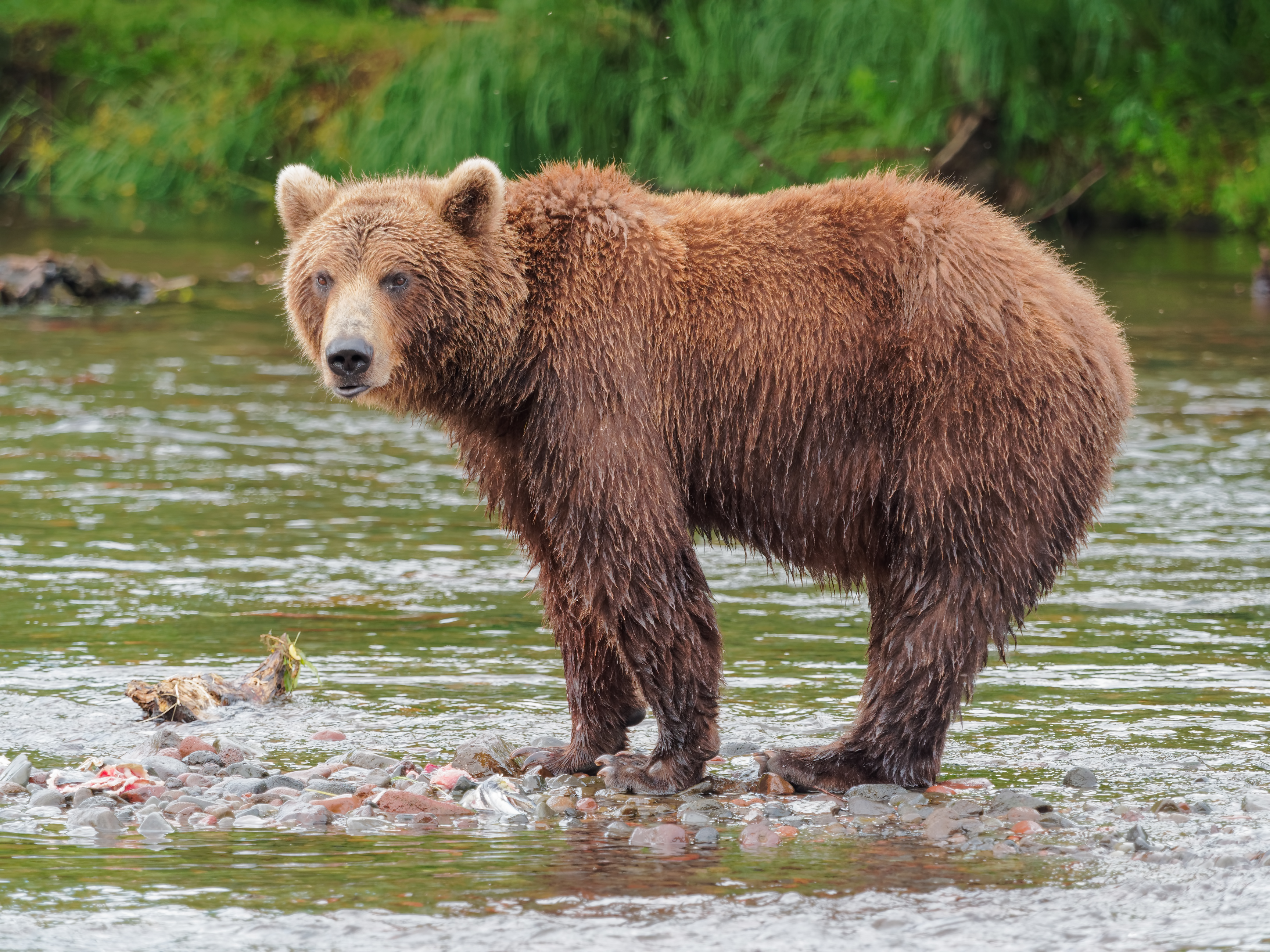}
\end{tabular}
&
\begin{tabular}{c}
\includegraphics[width=4.5cm]{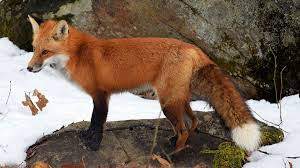}
\end{tabular}  \\
{\small \em Similar class: Brown bear}  &   {\small \em  Similar class: Red fox }
\end{tabular}
\ \\

\ \\ (c) {\small Class confusion}
\end{tabular}

\end{tabular}
\caption{\small Attention confusion and class confusion. (a): There is one main object in the image and it is visually distinct. Both the C-perplexity (1.26) and the X-perplexity (0.04) are low.
(b): There are multiple (two) main objects, which causes {\em attention confusion}.
The C-perplexity (1.77 ) and the X-perplexity (0.8) are higher than in the previous case.
(c): There is  one main object. However, it visually resembles objects from several other classes. This leads to {\em class confusion}.
The C-perplexity (2.3) and the X-perplexity (0.25) are higher than in the case of (a).}
\label{fig.confusion}
\end{figure}
\vspace{-1cm}
\end{center}

\begin{figure}[t]
\centering

%%%%%
\begin{tabular}{ccc}
\begin{tabular}{c}
\includegraphics[width=2.5cm]{figs/bird-mirror.JPEG}
\end{tabular}
&
\begin{tabular}{c}
\hspace{-0.5cm} \includegraphics[width=3cm]{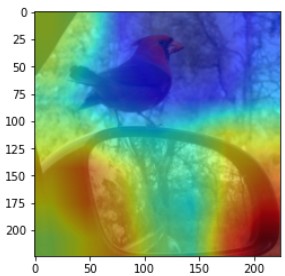}
\end{tabular}
&
\begin{tabular}{c}
\hspace{-0.5cm} \includegraphics[width=3.2cm,height=2.8cm]{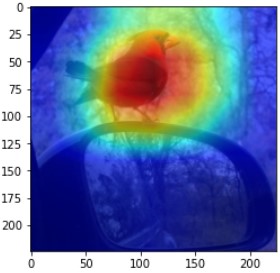}
\end{tabular}

\\
{\small Original}  &  \hspace{-0.3cm} {\small Car mirror }  &\hspace{-0.4cm} {\small House
flinch}
\end{tabular}

%%%%%
\begin{tabular}{cccc}
\begin{tabular}{c}
\includegraphics[width=4cm]{figs/class-confusion.JPEG}
\end{tabular}
&
\begin{tabular}{c}
\hspace{-1.2cm} \includegraphics[width=4.5cm]{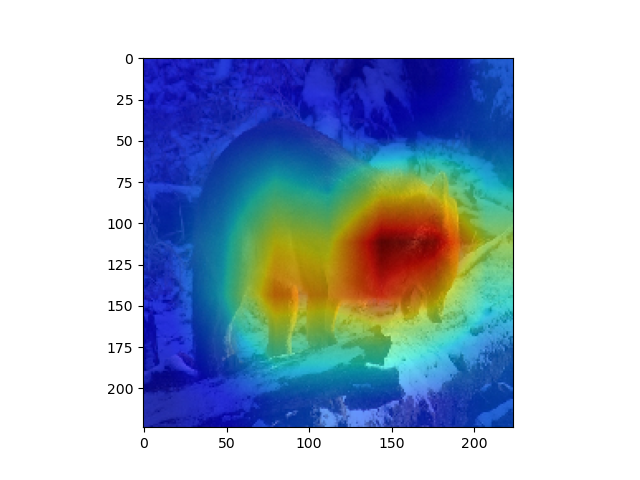}
\end{tabular}
&
\begin{tabular}{c}
\hspace{-1.5cm} \includegraphics[width=4.5cm]{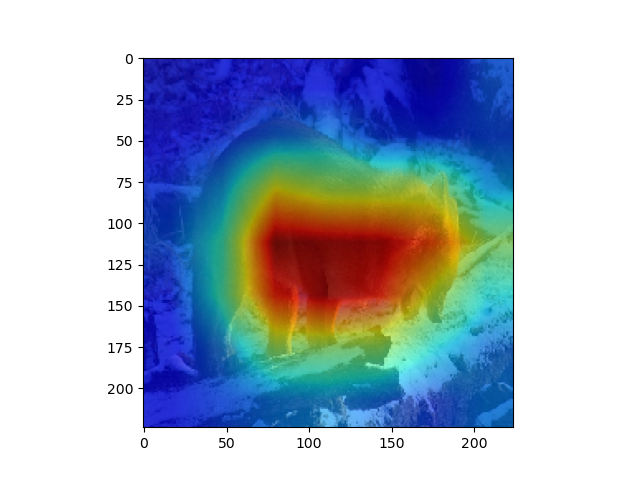}
\end{tabular}

&
\begin{tabular}{c}
\hspace{-1.5cm} \includegraphics[width=4.5cm]{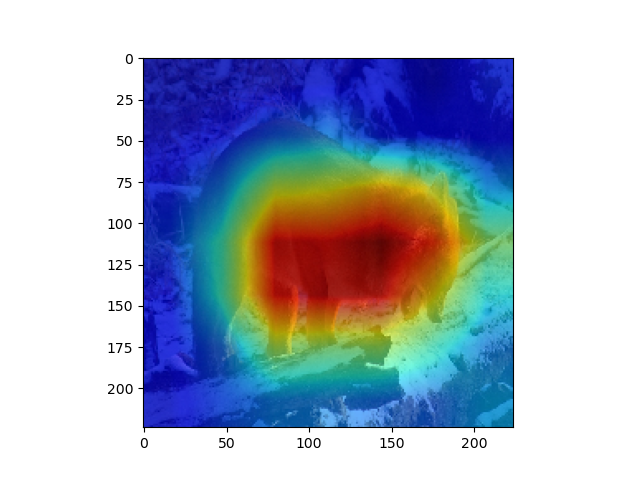}
\end{tabular}
\\
{\small Original}  &  \hspace{-1.2cm} {\small Dhole }  &\hspace{-1.5cm} {\small Brown bear}  & \hspace{-1.5cm} {\small Wild boar}
\end{tabular}

\caption{Grad-Cam heatmaps based on the best ResNet50 model. Top row: The classifier focuses at different regions of the image for different classes. The difficulty is to decide which part to focus on.
Bottom row: The classifier focuses on the same region for different classes. The difficulty is to decide what the object is.}
\label{grad-cam}
\end{figure}

How do we know if attention confusion and/or class confusion play an role in the classification of an example?  To answer this question, we need to introduce two concepts.
The fraction of the classifiers that classify an example $\x$ into a class $j$ is:

\[f(\x, j) = \frac{1}{N} \sum_{i=1}^N \1(C_i(\x) = j).\]

\noindent It can also be viewed as the fraction of the votes that $\x$ receives from the classifiers.
We call the class labels with the highest numbers of the votes the {\em top voted labels}
of $\x$.

The probability that $\x$ belongs to class $j$ according to classifier $i$ is $P_i(j|\x)$.
Imagine assigning $\x$ to a class via sampling according to $P_i(j|\x)$ rather than to assign it to the class with the maximum probability.  Then the expected number of classifiers that assign $\x$ to class $j$ would be:

\[f_e(\x, j) = \frac{1}{N} \sum_{i=1}^N P_i(j|\x).\]

\noindent
It can also be viewed as the expected fraction of the votes that $\x$ receives from the classifiers.
We call the class labels with the highest expected number votes the {\em top expected labels} of $\x$.

The top voted labels for the image in Figure \ref{fig.confusion} (b) are {\tt car mirror} and {\tt house flinch}. { The top expected labels are the same.}  Apparently, those class labels correspond to different regions of the image (Figure \ref{grad-cam}, tope row).  Therefore, we know that attention confusion is what makes the image difficult to classify for the classifiers.

For the image in Figure \ref{fig.confusion} (c) (top-right), the top 3 voted labels are {\tt dhole}, {\tt brown bear} and {\tt wild boar}.  The top 3 expected labels are {\tt dhole}, {\tt brown bear} and {\tt red fox}.
 All those classes are visually similar, and are referring to the same object in the image
 (Figure \ref{grad-cam}, bottom row).  Therefore, we know that class confusion is what makes the image difficult to classify for the classifiers.

\begin{center}
\begin{figure}[h!]
\bc

\begin{tabular}{c}
 \includegraphics[width=6cm]{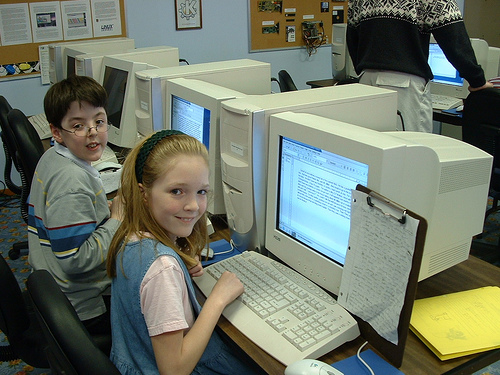}
\end{tabular}

\ec

%\vspace{-0.5cm}
\caption{\small Mixed confusion. The C-perplexity (5.7) and the X-perplexity (0.86)
are both high.
 }

\label{fig.confusion1}
\end{figure}
\vspace{-0.5cm}
\end{center}

Often, a classifier must confront  both attention confusion and class confusion. One example is shown in Figure \ref{fig.confusion1}.
 The ground truth label  of the image is {\tt screen}.  In the results of ResNet50, the five classes with the highest probability values are:
{\tt laptop} (0.58), {\tt notebook} (0.13), {\tt crossword puzzle} (0.07),
{\tt desk} (0.06), {\tt desktop computer} (0.03).
They  can be divided into three groups: (1) {\tt laptop}, {\tt notebook}, {\tt desktop computer}; (2) {\tt crossword puzzle}; and (3) {\tt desk}.  The difficulty in distinguishing between classes in the first group is due to class confusion, and the difficulty in distinguishing between different groups is due to attention confusion.

\section{Class Perplexity}
\label{class-diff}

The validation set of ImageNet consists of 1,000 classes, each with 50 images.
In the previous section, we have discussed perplexity of individual examples. In this section, we extend the concept to classes.
 Specifically,
the {\em C/X-perplexity of  a  class} is defined to be the average of the C/X-perplexity values of all the examples in the class.
\noindent   The classes with the highest C-perplexity and X-perplexity are given in Table \ref{top-class} .

\begin{table}[h!]
\centering
\caption{Classes with the highest C-perplexity and X-perplexity.}
\label{top-class}
\begin{tabular}{|l|l||l|l|}
\hline
 Class         & C-perplexity               & Class          & X-perplexity   \\ \hline
{\tt screwdriver}     & 7.78            & {\tt  velvet }       &  0.87   \\ \hline
{\tt loupe     }     & 7.04           & {\tt screen }       &   0.85       \\ \hline
{\tt miniskirt}      & 6.73           & {\tt  sunglass }     &   0.81    \\ \hline
{\tt sunglasses  }  & 6.48           &  {\tt water jug }     & 0.80   \\ \hline
{\tt hair spray }    & 6.17            & {\tt hook    }      & 0.77     \\ \hline
{\tt hatchet    }    & 6.15            & {\tt  spotlight }   & 0.77    \\ \hline
{\tt power drill }   & 6.10            & {\tt ladle    }     &0.76  \\ \hline
{\tt backpack   }    & 5.99            & {\tt English foxhound} & 0.75   \\ \hline
{\tt sunglass   }    & 5.77           & {\tt laptop   }      &  0.75    \\ \hline
{\tt plunger    }    & 5.61           & {\tt chiffonier}     & 0.75   \\ \hline
\end{tabular}
\end{table}

What factors contribute to high class perplexity?  In other words, what makes a class difficulty to classify?  To answer this question, we need to introduce some new concepts for classes.

As defined in the previous section, $f(\x, j)$ denotes the fraction of the votes that the example $\x$
receives from the classifiers.
 For a given class $c$,  let $f(c, j)$ be the average of
$f(\x, j)$ over all the examples in the class.
\noindent It is the frequency that class $c$ is confused as class $j$.
Note that $f(c, j) \neq f(j, c)$ in general.

  Define
the {\em voted-confusion} between the two classes to be $s(c, j) = \frac{f(c, j) + f(j, c)}{2}$.
 \noindent We refer to the class labels $j$ with highest $f(c, j)$ as the {\em top voted-confusion labels} of class $c$.  In a similar fashion,   we  define
 {\em expected-confusion} between classes and  {\em top expected-confusion labels} of a class.

\begin{figure}[h!]
\centering
\includegraphics[width=13cm,height=4cm]{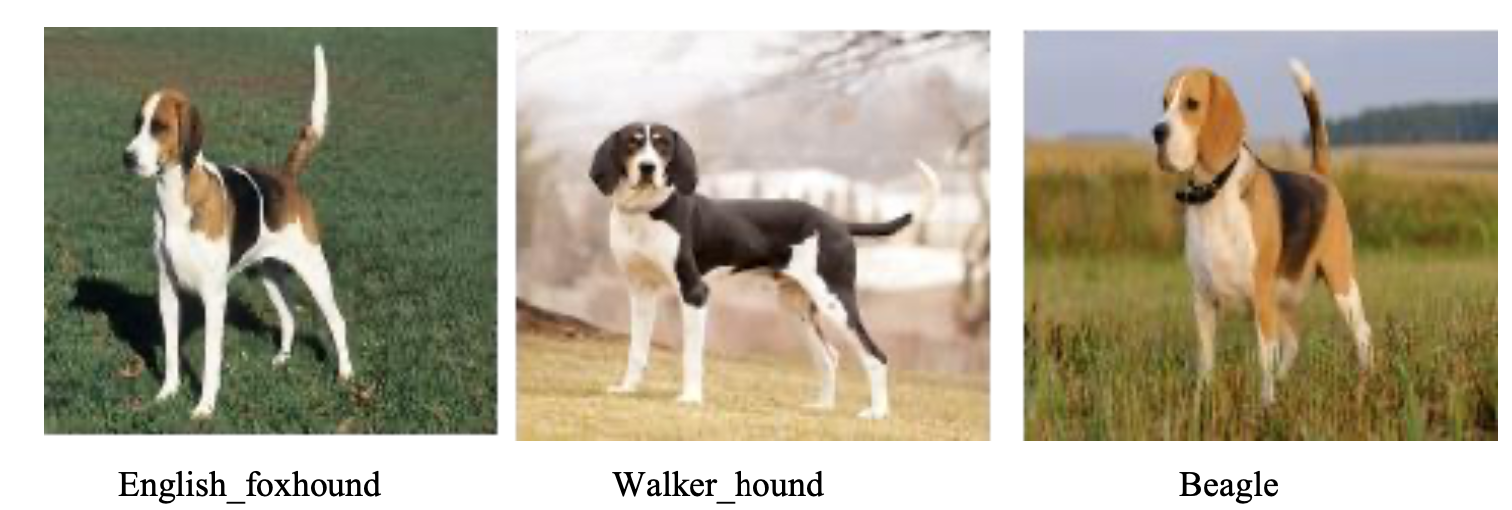}
\caption{Example images of the classes {\tt English foxhound}, {\tt Walker hound} and {\tt beagle} from ImageNet.  The three classes have high voted-confusion. }
\label{foxhound}
\vspace{-0.5cm}
\end{figure}

An inspection of pairs of top confusion classes reveals that two factors make a class difficult to classifier.  The first factor is {\em visual similarity between classes}. Take {\tt English foxhound} as an example.  It has high voted-confusion with {\tt Walker hound} and {\tt beagle}.  Examples of the three classes from ImageNet are  shown in Figure \ref{foxhound}. It is clear that the three classes do look very similar.

\bc
\ec

\begin{figure}[h!]
\centering
\includegraphics[width=13cm,height=4cm]{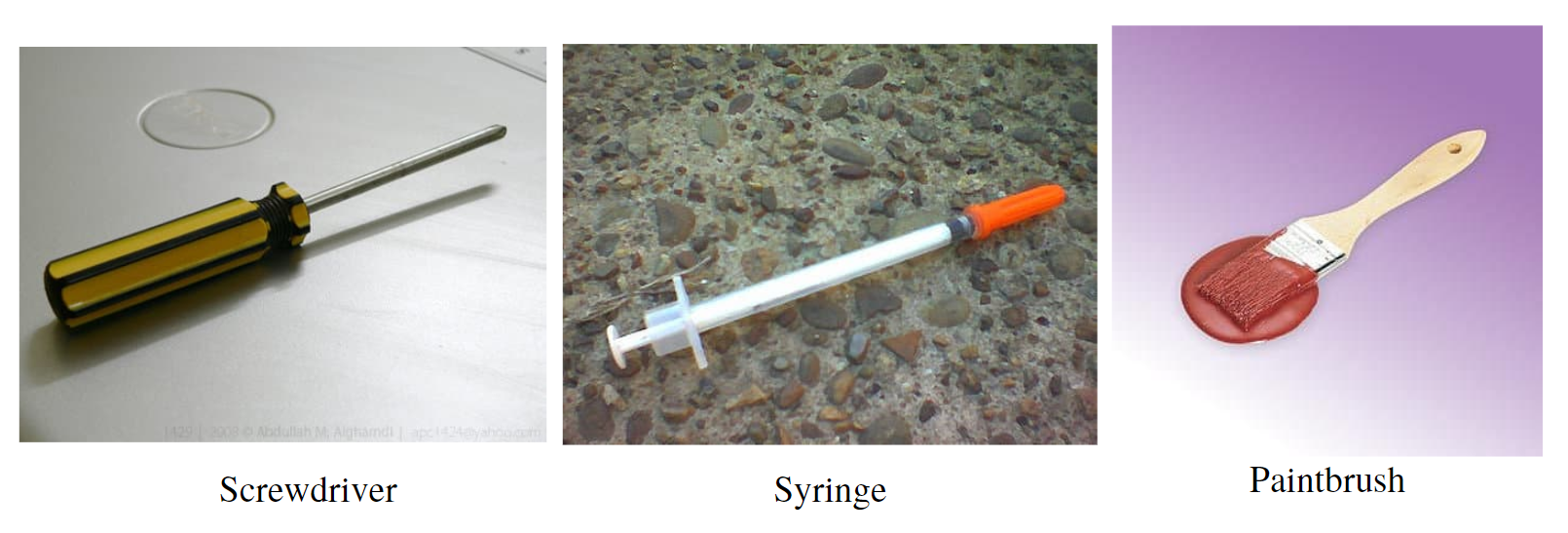}
\caption{Example images of the classes {\tt screwdriver}, {\tt syringe} and {\tt paintbrush} from ImageNet.  The three classes have high expected-confusion.}
\label{screwdriver}
\end{figure}

  The {\tt screwdriver} class has high expected-confusion with {\tt springe} and {\tt paintbrush}.  Examples of the three classes from ImageNet  are  shown in Figure \ref{screwdriver}. Although they might look very different for humans, they are actually very similar to  the classifiers.

The second reason for high class perplexity is {\em class co-occurence}.  For example,
the {\tt screen} class has high voted-confusion with {\tt desk} and {\tt mouse}.  Examples of the three classes from ImageNet  are  shown in Figure \ref{screen}.
In the image of each class, objects are the other two classes are also present.  As a matter of fact, those three classes co-occur frequently.

\begin{figure}[h!]
\centering
\includegraphics[width=13cm,height=4cm]{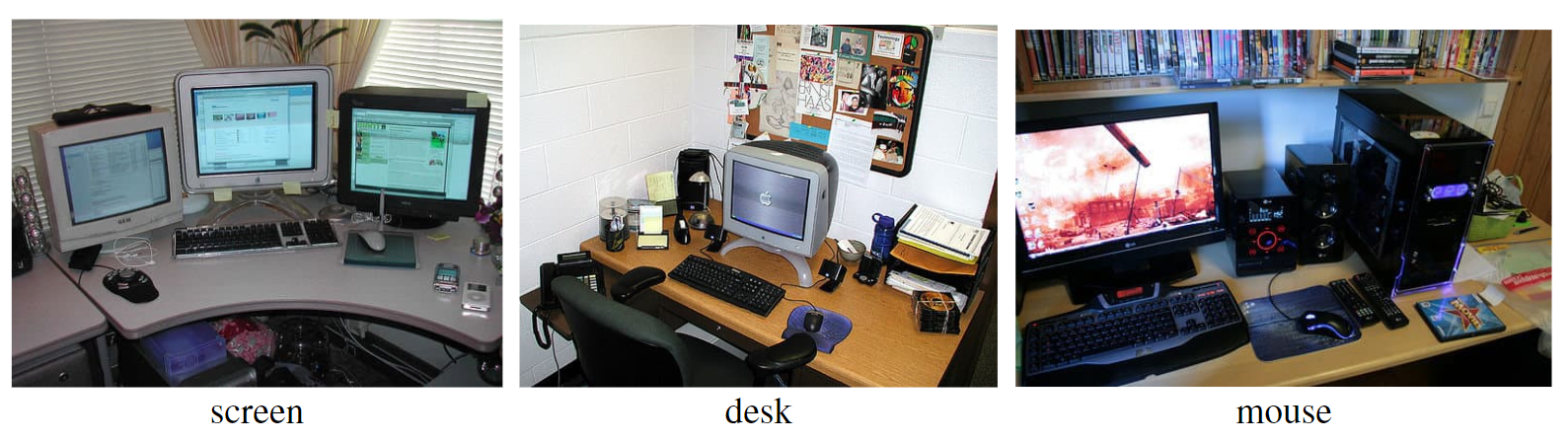}
\caption{Example images of the classes {\tt screen}, {\tt desk} and {\tt mouse} from ImageNet. The  three classes co-occur frequently.}
\label{screen}
\end{figure}

The {\tt velvet} class has high expected-confusion with {\tt studio couch} and {\tt pillow}.  Examples of the three classes from ImageNet  are  shown in Figure \ref{screen}. There is a strong co-occurrence relationship because the couch and the pillows are made of  materials that look like velvet.

\begin{figure}[h!]
\centering
\includegraphics[width=13cm,height=4cm]{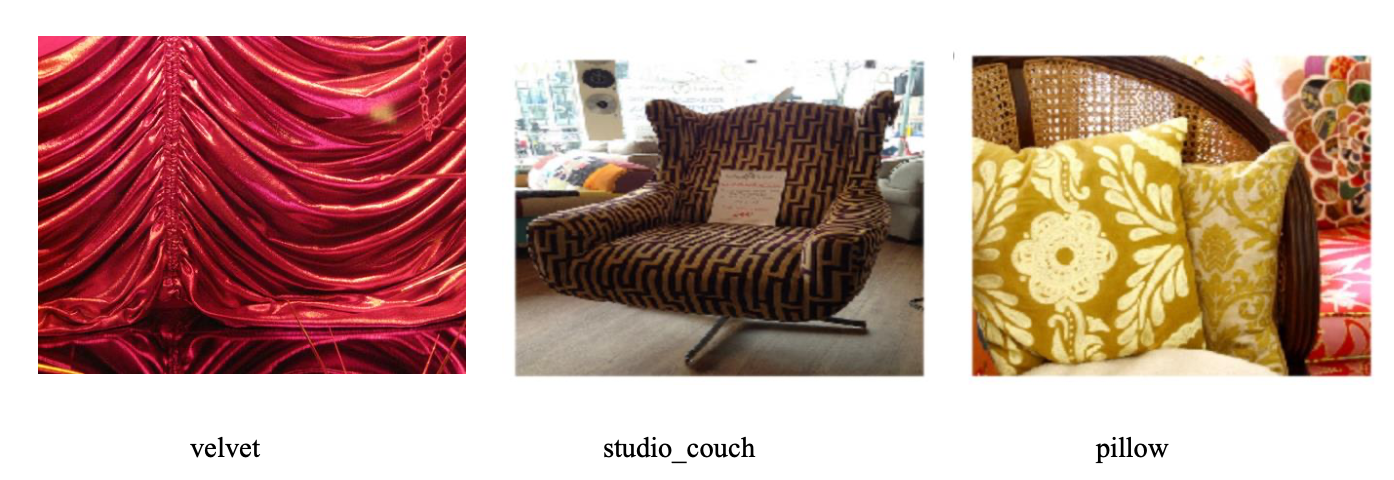}
\caption{Example images of the classes {\tt velvet}, {\tt studio couch} and {\tt pillow} from ImageNet. There is a strong co-occurrence relationship among the three classes because the couch and the pillows are made of  materials that look like velvet}
\label{screen}
\end{figure}

\section{Imperfections in the ImageNet Dataset}
\label{imperfections}

Out of curiosity, we have examined images with extremely high X-perplexity (close to 1), and we found that some examples in the ImageNet might have been incorrectly  or inappropriately labelled.  Figure  \ref{fig.mis-label} shows several examples.  They all have X-perplexity 1 and low C-perplexity.   In other words, they are classified incorrectly by the classifiers with high confidence.  As such, they are potentially mislabelled.

\begin{center}
\begin{figure}[h!]
\bc
\begin{tabular}{rcccc}
&
\begin{tabular}{c}
\includegraphics[width=2.5cm,height=2.5cm]{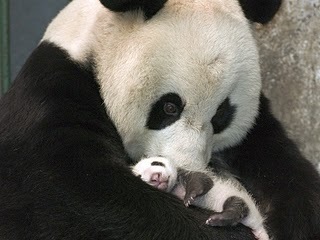}
\end{tabular}
&
\begin{tabular}{c}
\includegraphics[width=2.5cm,height=2.5cm]{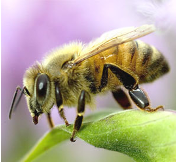}
\end{tabular}
&
\begin{tabular}{c}
 \includegraphics[width=2.5cm,height=2.5cm]{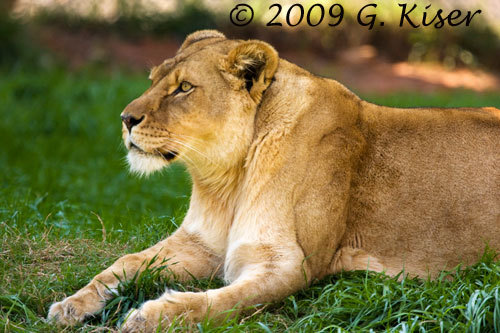}
\end{tabular}
&
\begin{tabular}{c}
 \includegraphics[width=2.5cm,height=2.5cm]{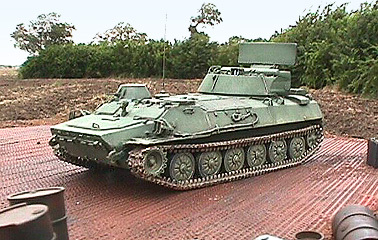}
\end{tabular}
  \ \\

& (a) & (b) &  (c)  &  (d) \\
ImageNet Label: & lesser panda & ant   & hussar monkey   & amphibian\\
X-Perplexity: &1 & 1   & 1   &1\\
C-Perplexity: & 1.00 &1.14   &   1.00 & 1.15\\
TVL: & giant panda &bee  & lion   &  tank\\

\end{tabular}
\ec

%\vspace{-0.5cm}
\caption{Potentially mislabeled images: ``TVL"  stands for top voted  label, i.e., the class label that receives the highest number of votes from the classifiers.}
\label{fig.mis-label}
\end{figure}
\vspace{-0.5cm}
\end{center}

Several other examples are shown in Figure \ref{fig.improper-label}.  Their  X-perplexity values  are also 1, while their C-perplexity values vary.  The top labels assigned by the classifiers seem to be more appropriate than the original ImageNet labels.  As such, those examples might have inappropriately labelled.

\begin{center}
\begin{figure}[h!]
\bc
\begin{tabular}{rccc}
&
\begin{tabular}{c}
\includegraphics[width=2.5cm,height=2.5cm]{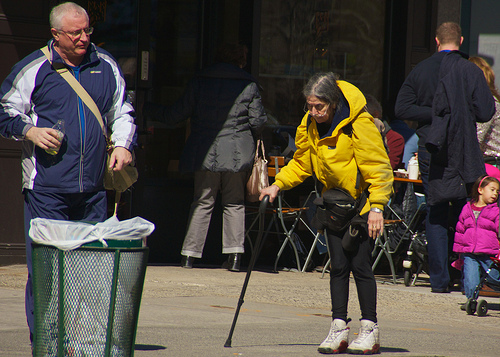}
\end{tabular}
&
\begin{tabular}{c}
\includegraphics[width=2.5cm,height=2.5cm]{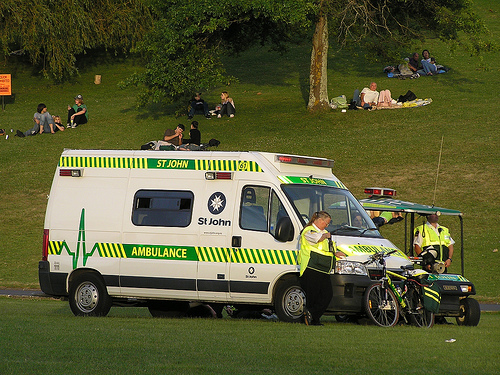}
\end{tabular}

&
\begin{tabular}{c}
 \includegraphics[width=2.5cm,height=2.5cm]{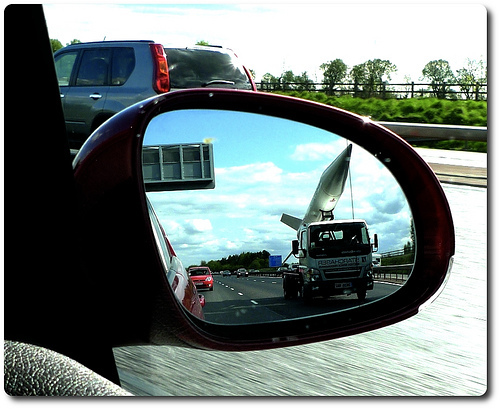}
\end{tabular}
  \ \\

& (a) & (b) &  (c)   \\
ImageNet Label: & digital watch & golfcart     &  missile\\
X-Perplexity: &1 & 1      &1\\
C-Perplexity: & 14.81&1.30   & 1.11\\
TVLs: & crutch, stretcher, sax &ambulance, police van  &  car mirror\\
TELs :& crutch, stretcher, sax &police van, ambulance &  car mirror, sunglasses\\

\end{tabular}
\ec

%\vspace{-0.5cm}
\caption{Potentially inappropriately labeled images.`TVLs"  stands for top voted  labels, i.e., the class labels that receive the most number of votes from the classifiers.
`TELs"  stands for top expected  labels,  i.e., the class labels that have the highest expected numbers of votes from the classifiers if class assignment is done by sampling.}
\label{fig.improper-label}
\end{figure}
\vspace{-0.5cm}
\end{center}

In Section \ref{class-diff}, we have highlighted two reasons for high class perplexity, visually similarity between classes and class co-occurrence.
An examination of highly confusing class pairs reveals  a third reason: Some class labels might have been inappropriately included in the dataset.
As shown in Figure \ref{fig.confusing-pairs}, both {\tt corn} and {\tt ear} are class labels in ImageNet. It is unclear how they differ from each other.   The same can be said about the  confusing pair  {\tt missile} and {\tt projectile}.

\begin{center}
\begin{figure}[htbp]
\bc
\begin{tabular}{rcccc}
&
\begin{tabular}{c}
\includegraphics[width=3cm,height=3cm]{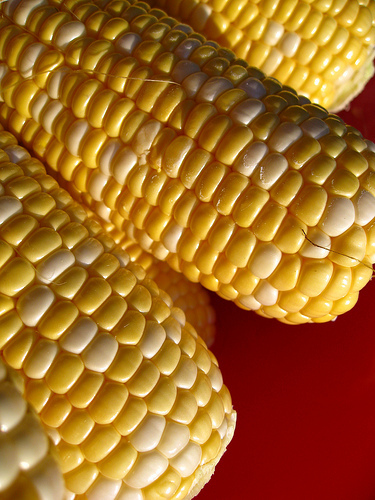}
\end{tabular}
&
\begin{tabular}{c}
 \includegraphics[width=3cm,height=3cm]{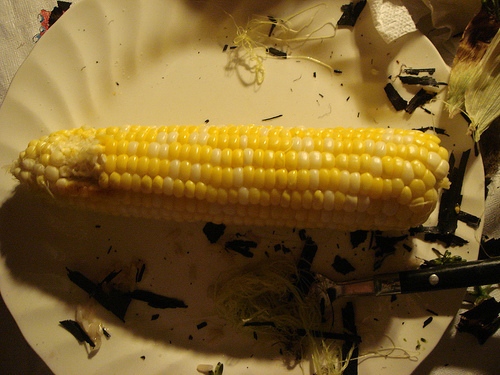}
\end{tabular}
&
\begin{tabular}{c}
 \includegraphics[width=3cm,height=3cm]{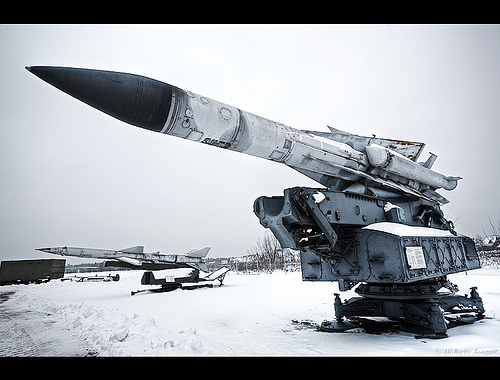}
\end{tabular}
&
\begin{tabular}{c}
 \includegraphics[width=3cm,height=3cm]{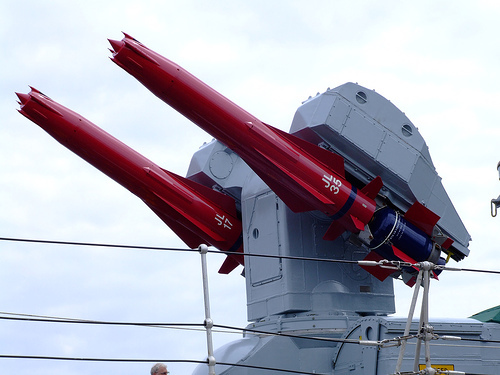}
\end{tabular}
  \ \\

  & Corn   & Ear   & Missile & Projectile \\

\end{tabular}
\ec

%\vspace{-0.5cm}
\caption{\small Potentially problematic class labels. }
\label{fig.confusing-pairs}
\end{figure}
\vspace{-0.5cm}
\end{center}

\section{Conclusions}

In this paper, we have mainly addressed two research questions:  (1) How to measure the perplexity of an example? (2) What factors contribute to high example perplexity?
We propose two ways to  measure example perplexity, namely C-perplexity and X-perplexity.  The theory and algorithm
for computing example perplexity are developed, and they are applied to ImageNet. The perplexity results on the ImageNet validation
set are analyzed. Several interesting findings are made. In particular,
some insights are gained regarding what makes some images harder to classify than others.

It is also found that perplexity analysis can reveal imperfections of a dataset, which can potentially help with data cleaning.  This seems to be an interesting direction for further research.

\subsection*{Acknowledgements}

Research on this paper was supported by Huawei Technologies Co., Ltd and Hong Kong Research Grants Council under grant 16204920. We also thank the deep learning computing framework MindSpore (\url{https://www.mindspore.cn}) and its team for the support on this work. 
%Research on this paper was supported by Huawei Technologies Co.,
%Ltd.\ under Project 19201840HWLB15Z015, and Hong
%Kong Research Grants Council under grant 16204920.

\section*{References}
%\bibliography{ref}

%\bibliography{mybibfile}
%

\bibliography{ref}

\begin{thebibliography}{1}

\bibitem{beyer2020we}
Lucas Beyer, Olivier~J H{\'e}naff, Alexander Kolesnikov, Xiaohua Zhai, and
  A{\"a}ron van~den Oord.
\newblock Are we done with imagenet?
\newblock {\em arXiv preprint arXiv:2006.07159}, 2020.

\bibitem{embretson2013item}
Susan~E Embretson and Steven~P Reise.
\newblock {\em Item response theory}.
\newblock Psychology Press, 2013.

\bibitem{li2017not}
Xiaoxiao Li, Ziwei Liu, Ping Luo, Chen Change~Loy, and Xiaoou Tang.
\newblock Not all pixels are equal: Difficulty-aware semantic segmentation via
  deep layer cascade.
\newblock In {\em Proceedings of the IEEE conference on computer vision and
  pattern recognition}, pages 3193--3202, 2017.

\bibitem{nie2019difficulty}
Dong Nie, Li~Wang, Lei Xiang, Sihang Zhou, Ehsan Adeli, and Dinggang Shen.
\newblock Difficulty-aware attention network with confidence learning for
  medical image segmentation.
\newblock In {\em Proceedings of the AAAI Conference on Artificial
  Intelligence}, volume~33, pages 1085--1092, 2019.

\bibitem{scheidegger2020efficient}
Florian Scheidegger, Roxana Istrate, Giovanni Mariani, Luca Benini, Costas
  Bekas, and Cristiano Malossi.
\newblock Efficient image dataset classification difficulty estimation for
  predicting deep-learning accuracy.
\newblock {\em The Visual Computer}, pages 1--18, 2020.

\bibitem{tsipras2020imagenet}
Dimitris Tsipras, Shibani Santurkar, Logan Engstrom, Andrew Ilyas, and
  Aleksander Madry.
\newblock From imagenet to image classification: Contextualizing progress on
  benchmarks.
\newblock {\em arXiv preprint arXiv:2005.11295}, 2020.

\bibitem{tudor2016hard}
Radu Tudor~Ionescu, Bogdan Alexe, Marius Leordeanu, Marius Popescu, Dim~P
  Papadopoulos, and Vittorio Ferrari.
\newblock How hard can it be? estimating the difficulty of visual search in an
  image.
\newblock In {\em Proceedings of the IEEE Conference on Computer Vision and
  Pattern Recognition}, pages 2157--2166, 2016.

\bibitem{yu2020difficulty}
Shuang Yu, Hong-Yu Zhou, Kai Ma, Cheng Bian, Chunyan Chu, Hanruo Liu, and
  Yefeng Zheng.
\newblock Difficulty-aware glaucoma classification with multi-rater consensus
  modeling.
\newblock {\em arXiv preprint arXiv:2007.14848}, 2020.

\end{thebibliography}

%\begin{btSect}{ref}
%\btPrintCited
%\end{btSect}

%{\small \bibliographystyle{splncs04}
%	\bibliography{ref} }

\newpage
\section*{Appendix 1: Correlation between C-Perplexity and X-Perplexity }

\begin{figure}[h]
	\centering
	\includegraphics[width=17cm, height=8cm]{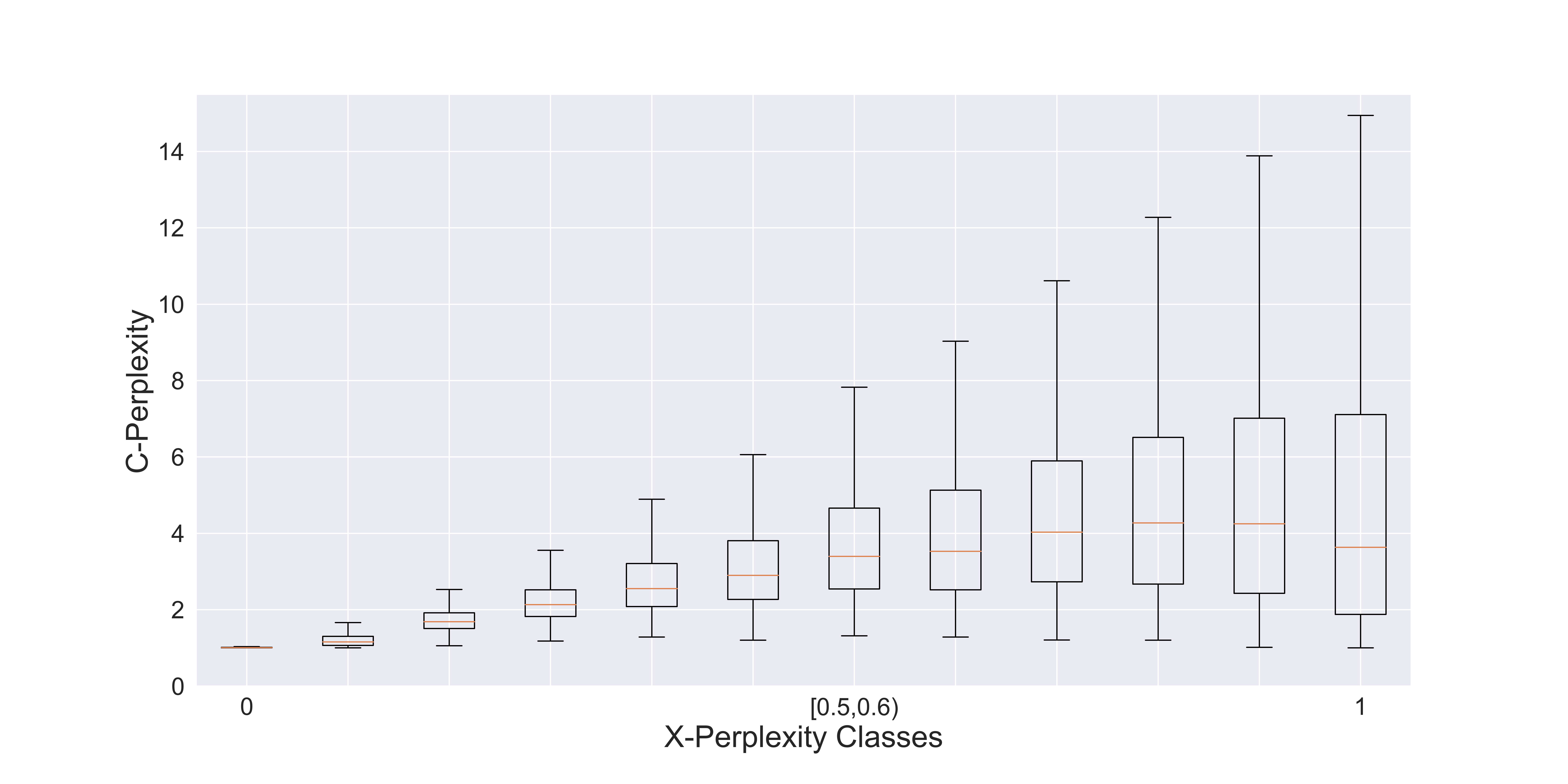}
	\caption{Box Chart of C-Perplexity corresponding to different X-Perplexity classes: looking the changing trend of 50\% and 75\% percentile line, generally with the increasing of X-Perplexity, the C-Perplexity increases also except the last X-Perplexity class.}
	\label{x_c_box}
\end{figure}

We can conclude the correlation relationship in one word: the X-Perplexity produced by the classifier population on ImageNet validation set is strongly correlated to C-Perplexity positively. The conclusion is suggested by Figure \ref{x_c_box}:  Box Chart of C-Perplexity corresponding to different X-Perplexity classes, the relatively high Pearson Correlation coefficient 0.63644 between X-Perplexity and C-Perplexity, and the high Spearman's Rank Correlation Coefficient\footnote{Spearman's rank correlation coefficient is a nonparametric measure of rank correlation (statistical dependence between the rankings of two variables).} 0.87425.

\newpage
\section*{Appendix 2:  Impact of Classifier Population on Perplexity }

Our classifier population includes  a wide range of  models with quite different prediction performance, starting from the best models trained by 100\% ImageNet training images to the weakest models trained by sampled 25\% training images. A remaining question here is how the perplexity is affected by the classifier population.

\

\begin{figure}[h!]
	\centering
	\includegraphics[width=10cm, height=6.5cm]{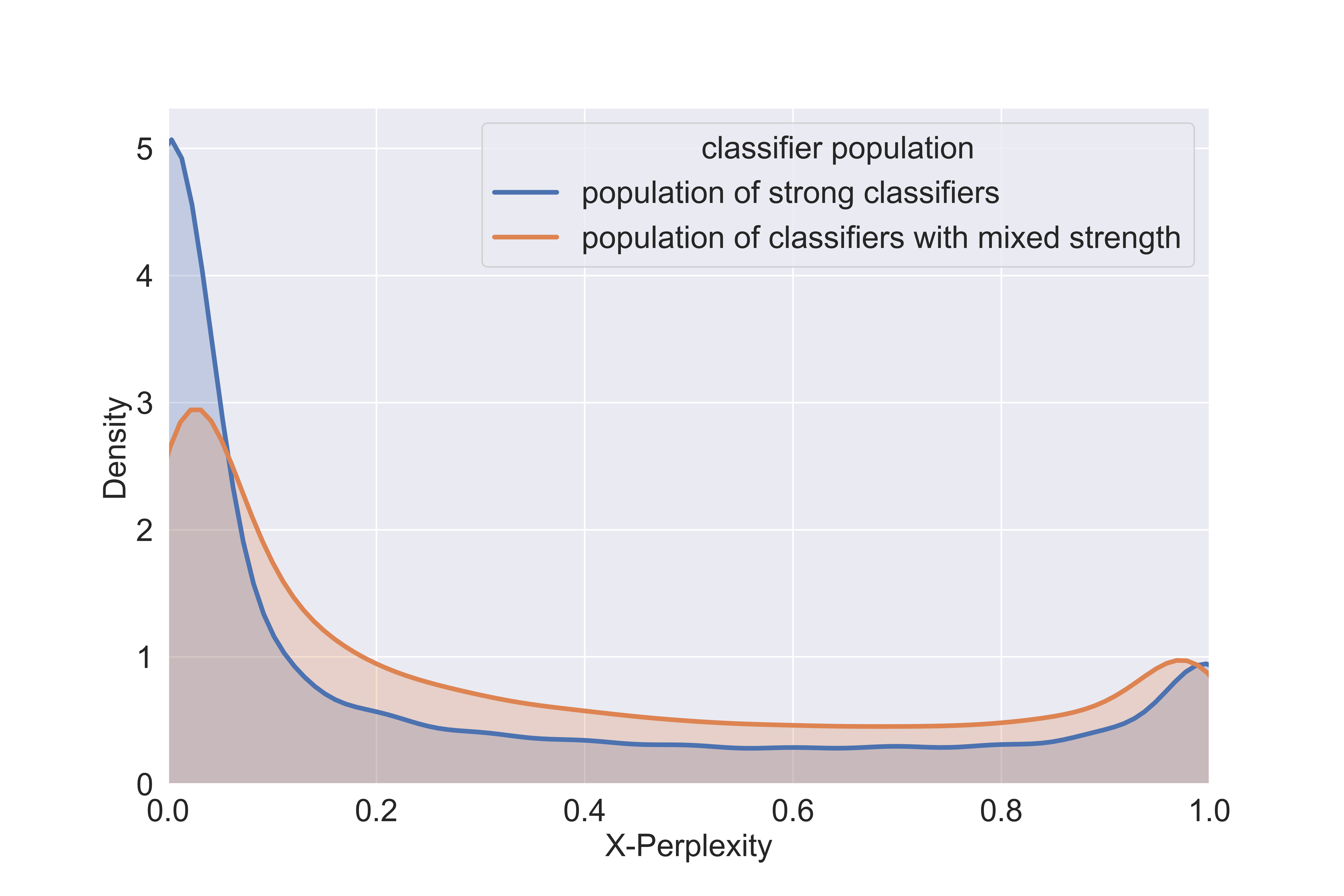}
	\includegraphics[width=10cm, height=6.5cm]{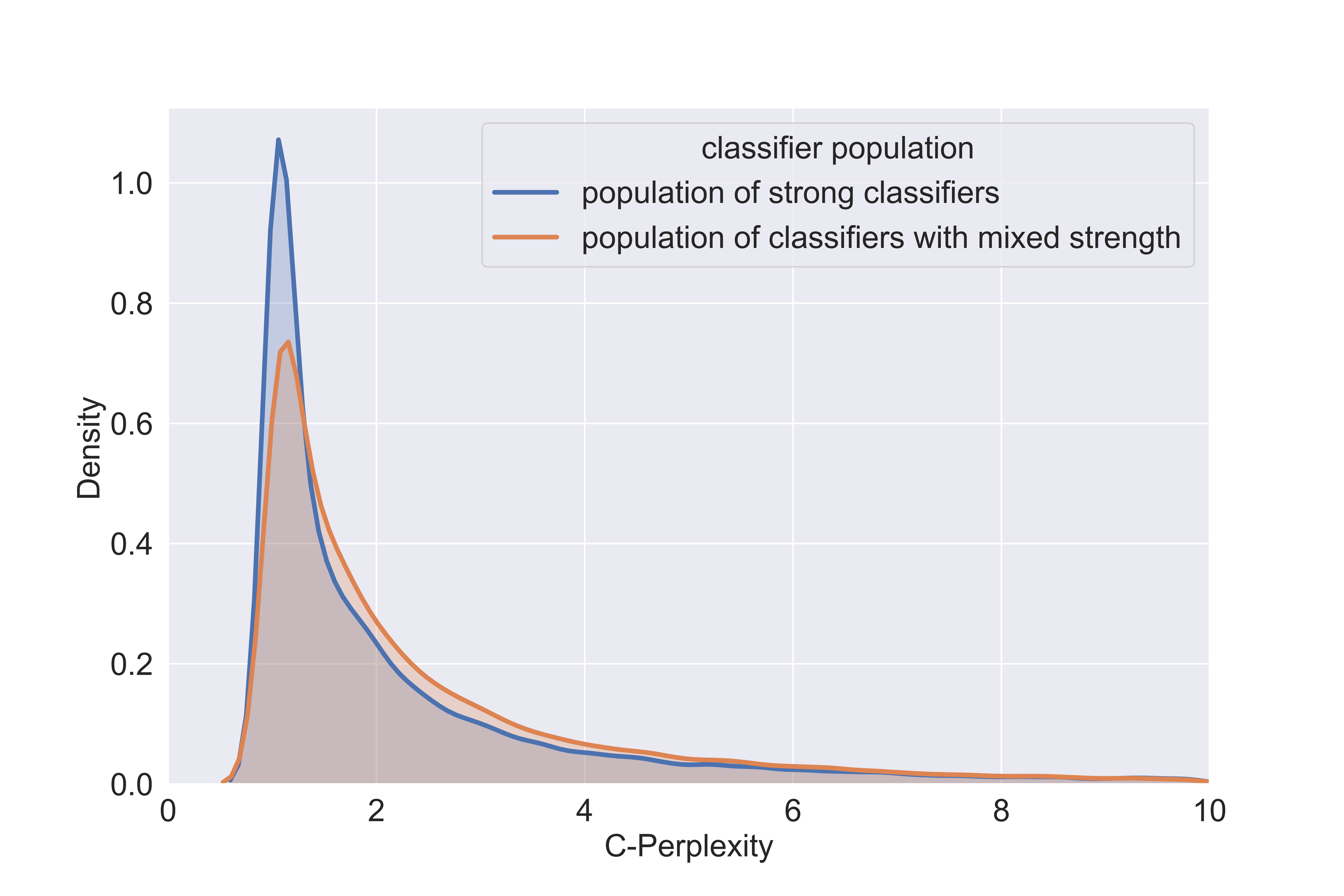}
	\caption{\scriptsize Comparison of X-Perplexity and C-Perplexity density plots (KDE) between population of strong classifiers (trained with 100\% ImageNet training images) and population of classifiers with mixed strength (the 500 classifiers). }
	\label{x_c_compare1}
\end{figure}

\begin{figure}[h!]
	\centering
	\includegraphics[width=10cm, height=6.5cm]{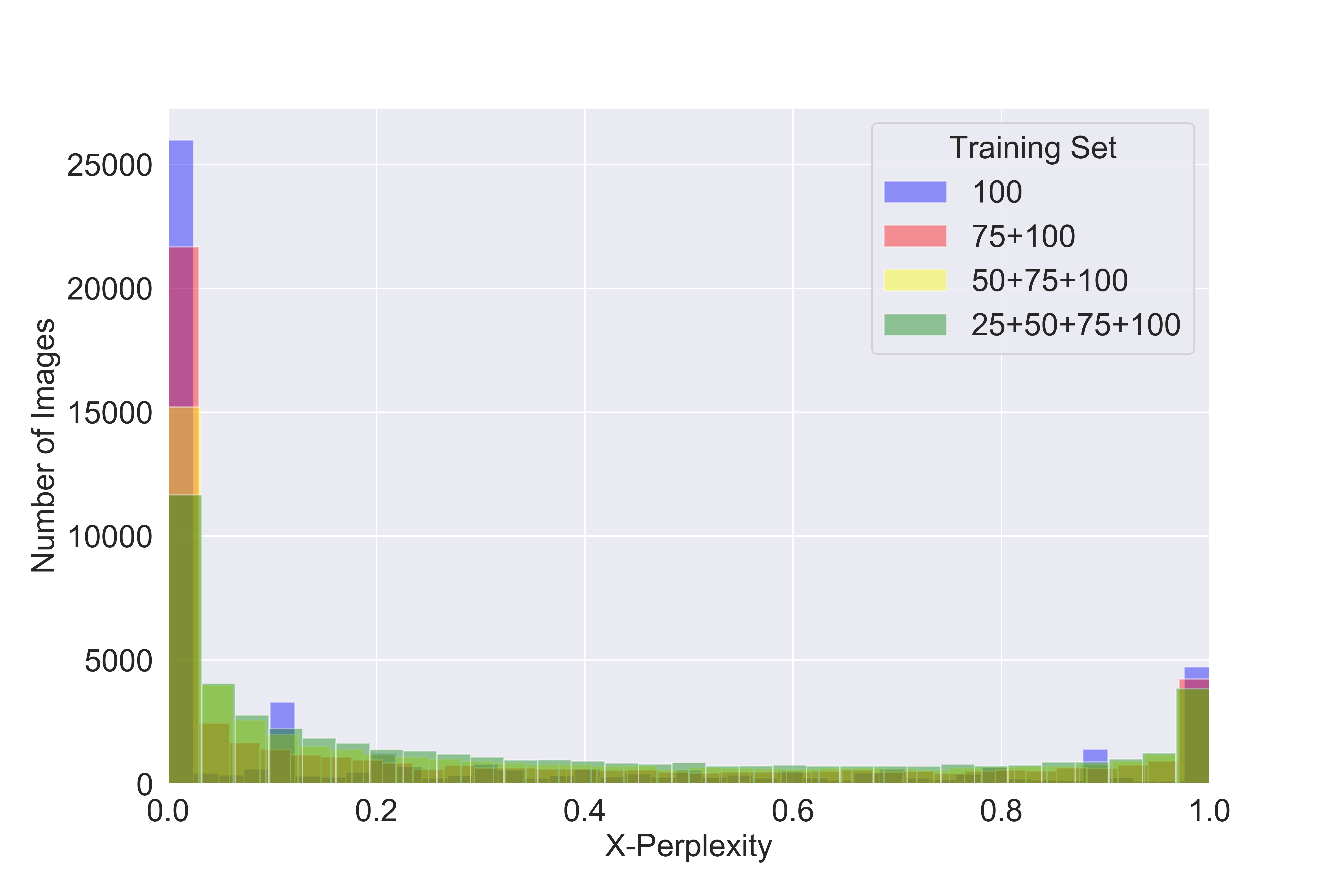}
	\includegraphics[width=10cm, height=6.5cm]{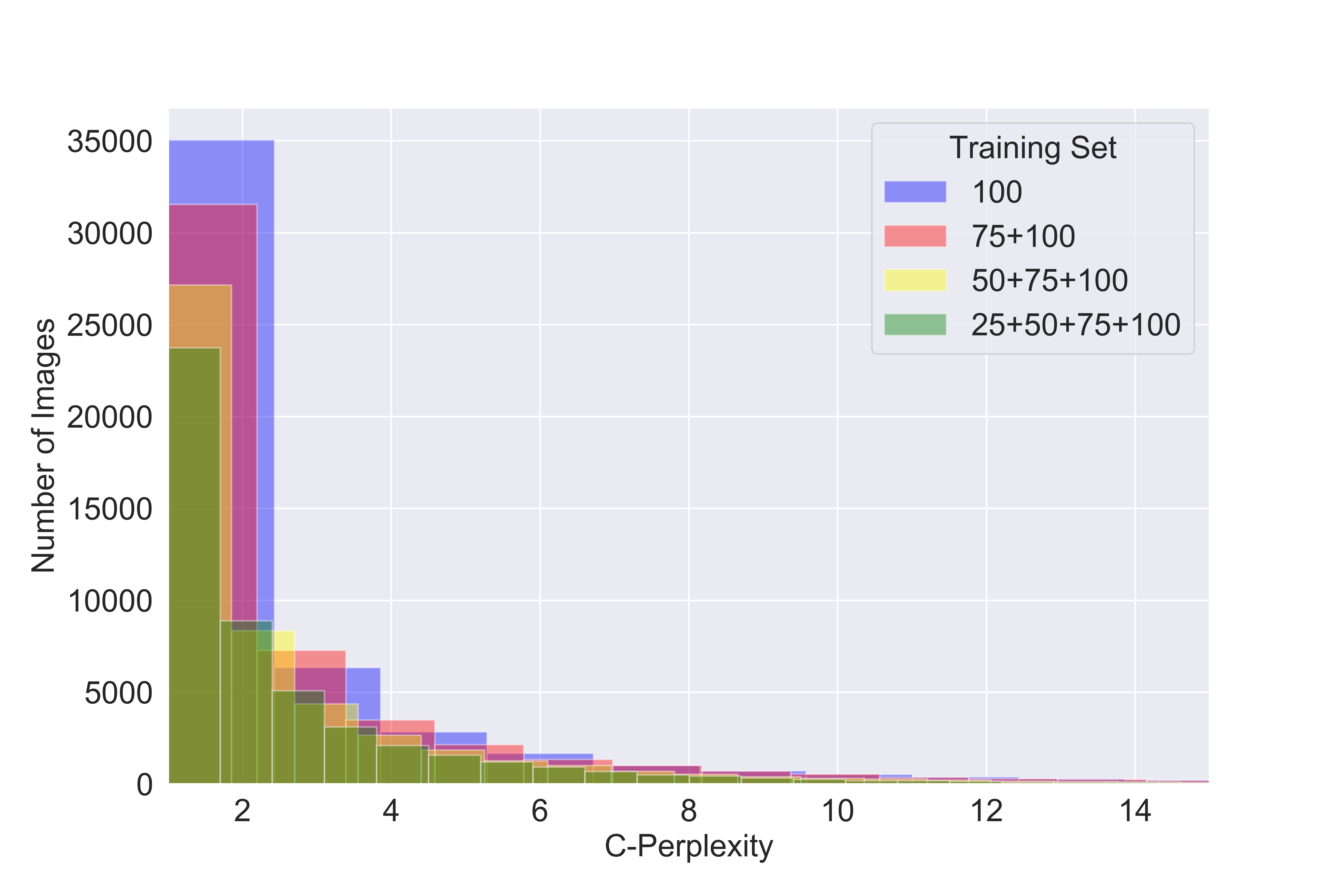}
	\caption{\scriptsize Comparison of X-Perplexity and C-Perplexity distribution with the adding of more weak classifiers: (1) 100-Population of best models trained by 100\% ImageNet training images; (2) 75+100-Add models trained by 75\%  ImageNet training images into the population; (3) 50+75+100-Add models trained by 50\%  ImageNet training images into the population; (4) 25+50+75+100-Add models trained by 25\%  ImageNet training images into the population.}
	\label{x_c_compare2}
\end{figure}

By the comparison of perplexity produced by different reference classifier population, we have two main findings:

a. With the increasing amount of weak classifiers in the reference classifier population, the more spread out the X-Perplexity and C-Perplexity distribution is. Figure \ref{x_c_compare1} shows that most of the images with  X-Perplexity closing to 0 and C-Perplexity closing to 1 under the population of strong classifiers, while the distribution under population of classifiers with mix strength is more dispersive. Similarly, from Figure \ref{x_c_compare2}, we can see that with more weaker classifiers adding into the population, the distribution become more and more spread out.

b. Although perplexity values vary according to the reference classifier population, the ordering they induce among the examples is stable across of a wide range of classifier populations generally. Specifically, the correlations are strong between populations with similar mixture of strong and weak classifiers. The correlations are relatively weak if the two populations have very different mixtures of strong and weak classifiers, for example one population consists of only strong classifiers and the other contains a significant fraction of weak classifiers.

\begin{figure}[htb]
	\centering
	\includegraphics[width=7cm, height=7cm]{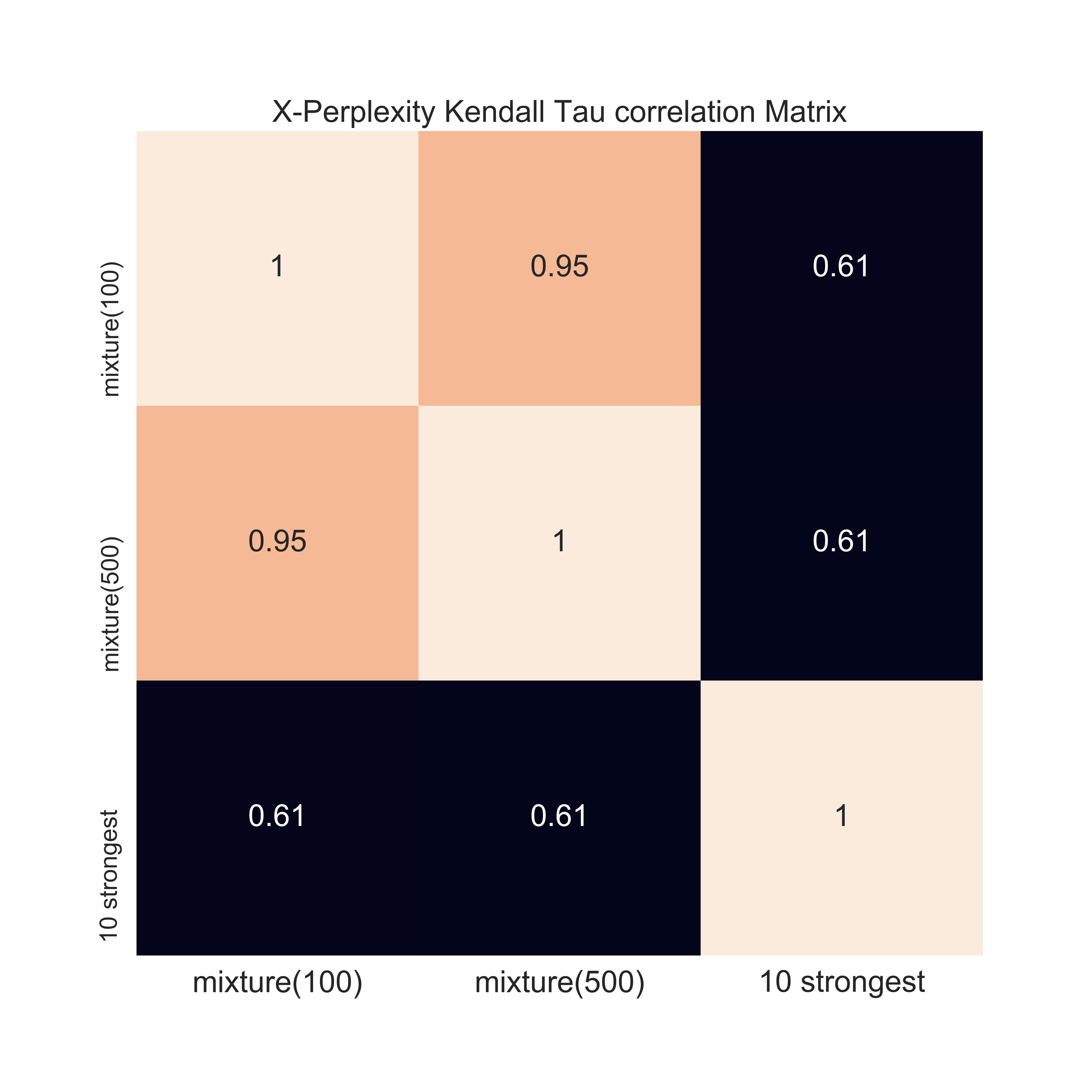}
	\includegraphics[width=7cm, height=7cm]{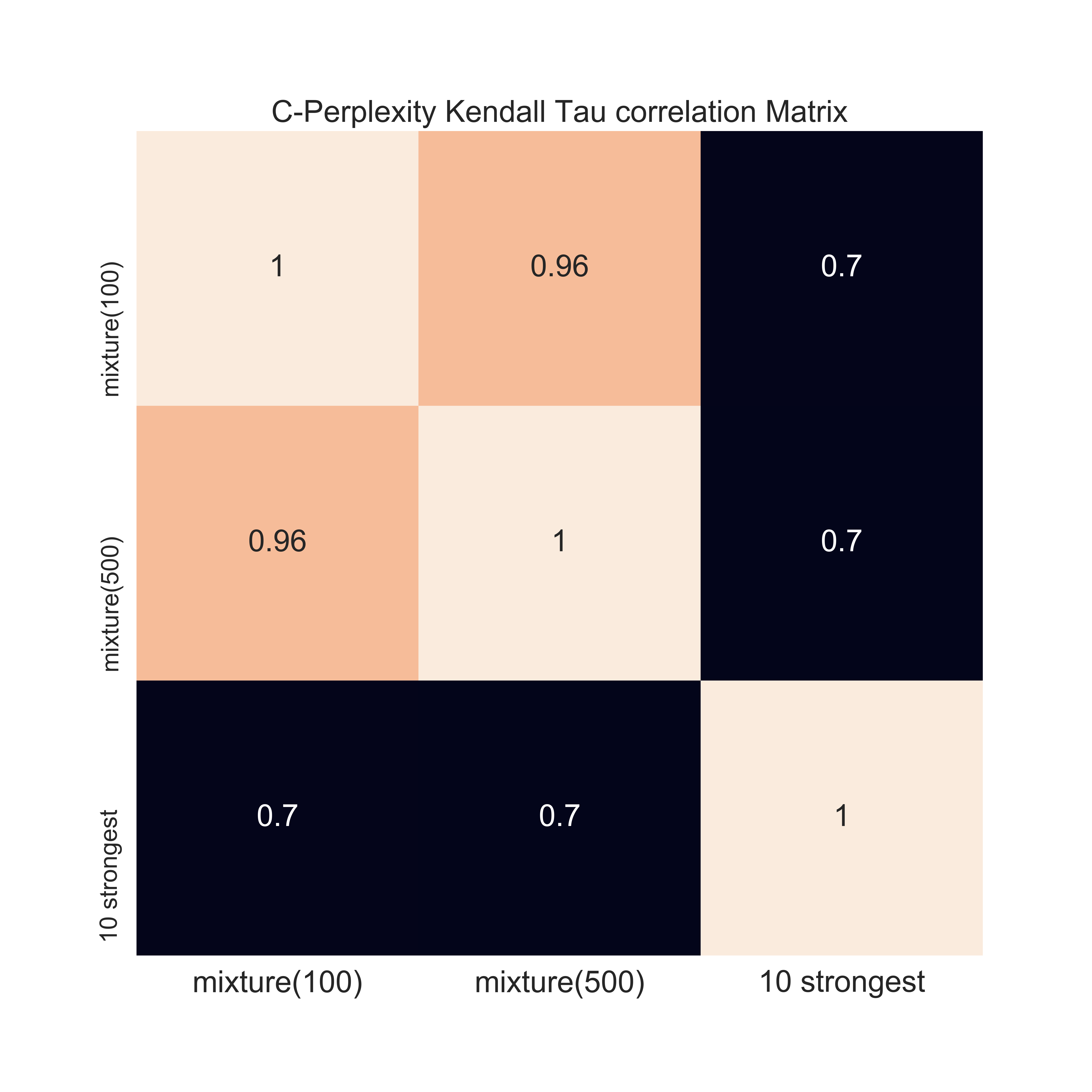}
	\label{x_c_compare3}
	\begin{center}
		\scriptsize

		Mixture(100)- 25\% ImageNet training images training model (30)+50\% ImageNet training images training model (30)+75\% ImageNet training images training model (30)+100\% ImageNet training images training model (10); \\
		
		\
		
		Mixture(500)- 25\% ImageNet training images training model (300)+50\% ImageNet training images training model (300)+75\% ImageNet training images training model (300)+100\% ImageNet training images training model (100); \\
		
		\
		
		10 strongest: Only include 10 Best models trained by  100\% ImageNet training images.
		
	\end{center}
	\caption{Kendall Tau Correlation matrix for X-Perplexity and C-Perplexity under different classifier population}
\end{figure}

The Kendall Tau correlation \footnote{Kendall rank correlation coefficient $\tau=\frac{\textnormal{number of concordant pairs}-\textnormal{number of discordant pairs}}{ \tbinom{n}{2}}$. It is a statistic used to measure the ordinal association between two measured quantities.} matrix in Figure \ref{x_c_compare3} for X-Perplexity and C-Perplexity show the evidence for this point. For population of  ‘Mixture(100)’ and ‘Mixture(500)’ , the mixture of strong and weak classifiers in these two population is similar and we can witness the high Kendall Tau correlation coefficient for both X-Perplexity(0.95) and C-Perplexity(0.96)  between these two population. In contrast, the Kendall Tau correlation coefficient between ‘Mixture(100)’ and ‘10 strongest’ is much smaller, which is 0.61 for X-Perplexity and 0.7 for C-Perplexity.

This is another justification for the use of a population of classifiers with various strengths, since the population including mixture of weak and strong classifiers is robust enough for practical purposes while the classifiers including only strong classifiers shows lower  robustness.

\newpage

\end{document}